\documentclass[11pt,a4paper]{article}
\usepackage[hyperref]{acl2021}
\usepackage{times}
\usepackage{latexsym}

\usepackage{microtype}

\aclfinalcopy % Uncomment this line for the final submission
\usepackage{multirow}

\usepackage{subfigure,graphicx}
\usepackage{algorithm}  
\usepackage{algpseudocode}  
\usepackage{amsmath}  

\usepackage{stfloats}
\usepackage{enumerate}

\usepackage{paralist}

\title{Using Adversarial Attacks to Reveal the Statistical Bias in Machine Reading Comprehension Models}

\author{Jieyu Lin\textnormal{\textsuperscript{2}},
 Jiajie Zou\textnormal{\textsuperscript{2}},
 Nai Ding\textnormal{\textsuperscript{1,2}\Thanks{    Corresponding author: Nai Ding}}\\
\textsuperscript{1}Zhejiang Lab / Hangzhou, China \\
\textsuperscript{2}Key Laboratory for Biomedical Engineering of Ministry of Education, College of \\Biomedical Engineering and Instrument Sciences, Zhejiang University / Hangzhou,\\ China \\
  \texttt{\{ljy5905,jiajiezou,ding\_nai\}@zju.edu.cn} \\}

\date{}

\begin{document}

\maketitle
\begin{abstract}
    Pre-trained language models have achieved human-level performance on many Machine Reading Comprehension (MRC) tasks, but it remains unclear whether these models truly understand language or answer questions by exploiting statistical biases in datasets. Here, we demonstrate a simple yet effective method to attack MRC models and reveal the statistical biases in these models. We apply the method to the RACE dataset, for which the answer to each MRC question is selected from 4 options. It is found that several pre-trained language models, including BERT, ALBERT, and RoBERTa, show consistent preference to some options, even when these options are irrelevant to the question. When interfered by these irrelevant options, the performance of MRC models can be reduced from human-level performance to the chance-level performance. Human readers, however, are not clearly affected by these irrelevant options. Finally, we propose an augmented training method that can greatly reduce models' statistical biases.
\end{abstract}

\section{Introduction}
\label{section1}
Reading comprehension tasks are useful to quantify language ability of both humans and machines
\citep{richardson2013mctest, xie-etal-2018-large,berzak-etal-2020-starc}. 
Deep neural network (DNN) models have achieved high
performance on many MRC tasks, but these models are not easily
explainable \citep{devlin-etal-2019-bert, brown2020language}. 
It is also shown that DNN models are often sensitive to adversarial attacks 
\citep{jia-liang-2017-adversarial,ribeiro2018semantically,si2019does,si2020benchmarking}. 
Furthermore, it has been shown DNN models can solve MRC tasks with
relatively high accuracy when crucial information is removed so that the
tasks are no longer solvable by humans
\citep{gururangan-etal-2018-annotation,si2019does, berzak-etal-2020-starc}.
All such evidence
suggests that the high accuracy DNN models achieve on MRC tasks does not
solely rely on these models' language comprehension ability. At least to
some extent, the high accuracy reflects exploitation of statistical
biases in the datasets 
\citep{gururangan-etal-2018-annotation,si2019does,berzak-etal-2020-starc}.

Here, we propose a new model-independent method to evaluate to what
extent models solve MRC tasks by exploiting statistical biases in the
dataset. As a case study, we only focus on the classic RACE dataset 
\citep{lai-etal-2017-race}, 
which requires MRC models to answer multiple-choice
reading comprehension questions based on a passage. The advantage of multiple-choice
questions is that its performance can be objectively evaluated. At the
same time, it does not require the answer to be within the passage,
allowing to test, e.g., the summarization or inference ability of
models. Nevertheless, since models are trained to select the right option from 4 options, which are designed by humans and may contain
statistical biases, models may learn statistical properties of the right option. Consequently, models may tend to select options with these
statistical properties similar to the properties of the right option without referring to the passage and question. Our method is designed to
reveal this kind of statistical bias.

The logic of our method is straightforward: For each multiple-choice
question, we gather a large number of options that are irrelevant to the
question and passage. We ask the model to score how likely each
irrelevant option is the right option. If a model is biased, it may
always assign higher scores to some irrelevant options than others, even if all the options are irrelevant. If a model
is so severely biased, which turns out to be true for all models
tested here, it may assign higher scores to some irrelevant options than
the true answer and select the irrelevant option as the answer. Here, the irrelevant options that are often selected as the answer was referred to as magnet options.

\section{Dataset and Pre-trained Models}
\label{section2}
We used RACE dataset in our experiment \citep{lai-etal-2017-race}, which is a
large-scale reading comprehension data set covering more than 28,000
passages and nearly 100,000 questions. The task was to answer
multi-choice questions based on a passage. Specifically, each question
contained a triplet (\(p_{i}\), \(q_{i}\), \(o_{i}\)), where \(p_{i}\)
denoted a passage, \(q_{i}\) denoted a question, and \(o_{i}\) denoted a
candidate set of 4 options, i.e.,
\(o_{i} = \{ o_{i,1},o_{i,2},o_{i,3},o_{i,4}\}\). Only one option was
the correct answer, and the accuracy was evaluated by the percent of
questions being correctly answered.

We tested 3 pre-trained language models, i.e., BERT 
\citep{devlin-etal-2019-bert}, 
RoBERTa \citep{liu2019roberta}, and 
ALBERT \citep{lan2019albert}. 
For each model, we separately tested the base
version and large version. We built our models based on pre-trained
transformer models in the Huggingface \citep{wolf2019huggingface}. 
We fine-tuned pre-trained models based on
the RACE dataset and the parameters we used for fine-tuning were shown
in Appendix \ref{a-1}.

The passage, question, and an option were concatenated as the input to models, i.e., {[}$CLS$, \(p_{i}\), $SEP$, \(q_{i}\),
\(o_{i,j}\), $SEP${]}. The 4 options were separately encoded. The
concatenated sequence was encoded through the models and the
output embedding of $CLS$ was denoted as \(C_{i,j}\). We used a linear transformation to convert vector \(C_{i,j}\) into a scalar $S(o_{i,j})$, i.e.,
\(S(o_{i,j}) = WC_{i,j}\). The scalar \(S(o_{i,j})\) was referred to as
the score of the option \(o_{i,j}\). A score was calculated for each
option, and the answer to a question was determined as the option with
the highest score, i.e., $\text{argmax}_{j}{S(o_{i,j})}$.

\section{Adversarial Method}
\label{section3}
\subsection{Screen Potential Magnet Options}
\label{section3.1}
We evaluated potential statistical biases in a model by giving it a large number of irrelevant options. For each question, we
augmented the options using a set of irrelevant options, i.e.,
\(O_{A} = \{ o_{a1},o_{a2},...,o_{aN}\}\). \(O_{A}\) was randomly selected
from the RACE dataset with 2 constraints. First, the options belonged to
questions that were not targeted at passage \(p_{i}\). Second, none of
the options in \(O_{A}\) was identical to any of the original options in
\(o_{i}\). The augmented question was denoted as
\((p_{i},q_{i},\{ o_{i,1},o_{i,2},o_{i,3},o_{i,4},o_{a1},...,o_{aj},...,o_{aN}\})\).
A score was independently computed for each option using the procedure mentioned above. Since the options in \(O_{A}\) were irrelevant, an ideal
model should never select them as answers. If
$\text{max}_{j}{S(o_{i,j})} < S(o_{ak})$ for any \(k\),
however, the model would select the \(k^{\text{th}}\) irrelevant option
as the answer. We define an interference score \(T_{k}\) using the
following equation.
% begin to write equation
\begin{equation*}
    T_{k} = \frac{1}{N}\sum_{i=1}^{N} T_{i,k},\qquad where
\end{equation*}
\begin{equation*}
    T_{i,k}=\begin{cases}
    1, \qquad if  \quad \text{max}_{j}{S(o_{i,j})} < S(o_{ak})\\
    0, \qquad otherwise
    \end{cases}
\end{equation*}
For an ideal model, \(T_{i,k}\) should always be 0. For a model that
makes mistakes but shows no consistent bias, the interference score
should be comparable for all \(o_{ak}\). If the model is biased,
the interference score may be always high for some options so that the
model always selects them as the answer whether they are relevant to the
question or not.

\subsection{Adversarial Attack}
\label{section3.2}
\begin{figure*}
    \begin{center}
        \begin{minipage}[tbp]{0.9\linewidth}
            \includegraphics[width=1\linewidth]{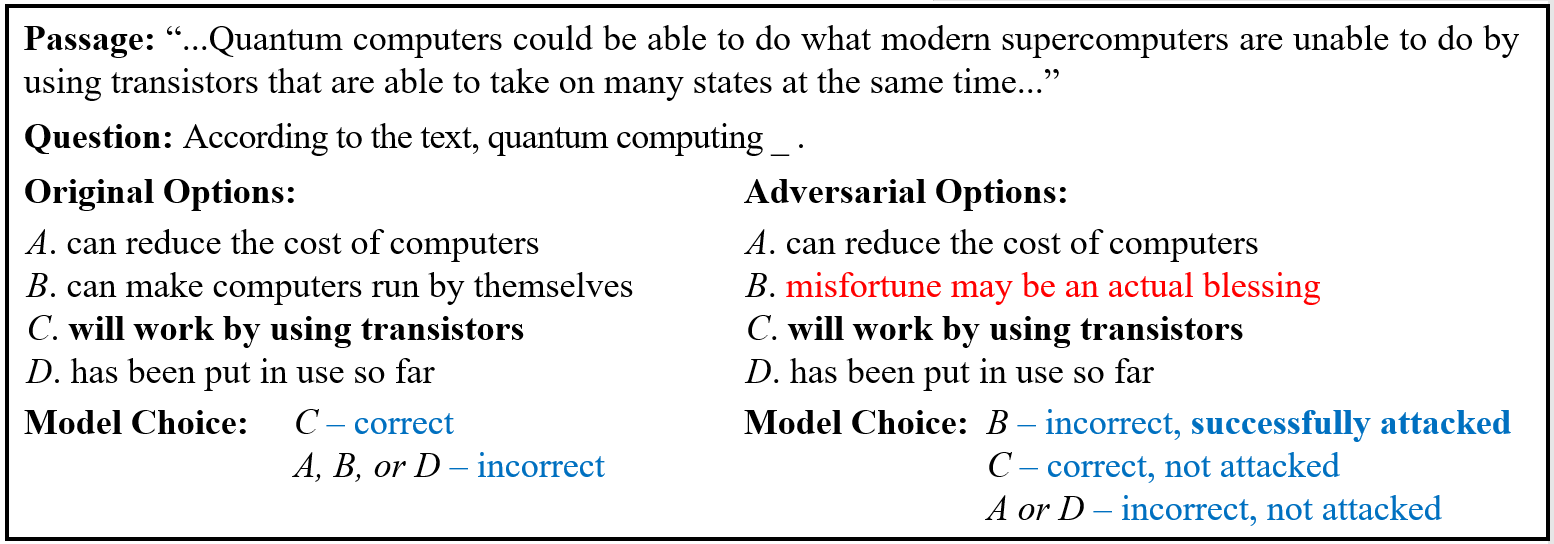}
        \end{minipage}%
        
    \end{center}
    \caption{\label{fig3.2} 
    An example of the task and adversarial attack. The option in bold is the true answer, and the option in red indicates the irrelevant option that was used for attack. 
    }
\end{figure*}
We constructed an adversary attack to the MRC models using one magnet option. For each question, we replaced a wrong option with a magnet option, i.e., $o_{ak}$. The replaced option set was $\{o_{i,1}, o_{i,2}, o_{i,3}, o_{ak}\}$. The passage and the question were not modified, and the answer did not change. An example was shown in Figure \ref{fig3.2}. If the model chooses the original answer even when a magnet option is introduced, it is stable, not sensitive to the attack. In contrast, if it chooses the magnet option, i.e., $o_{ak}$, as the answer, it is successfully attacked.

\section{Results and Analysis}
\label{section4}
\subsection{Experiments Setup}
\label{section4.1}
To screen potential magnet options, we constructed a large set of
irrelevant options, i.e., \(O_{A}\), by randomly selecting  300
passages from the RACE test set, which were associated with 1064 questions.
Furthermore, to test whether options in the training set can cause
stronger interference, we also randomly selected 300 passages from the
RACE training set, which had 1029 questions. The options from
the test and training set were pooled to create \(O_{A}\),
which had 8372 options in total.

For such a large number of irrelevant options, it was computationally challenging
to evaluate the interference score of each option based on each question
in the RACE test set. Therefore, as a screening procedure, we first
randomly selected 100 passages from the RACE test set, which have a
total of 346 questions. The interference score for each of the 8372 irrelevant
options was evaluated based on the 346 questions. 

After potential magnet options were determined by the screening
procedure, the interference score of magnet options were further
evaluated using all questions in RACE test set. For RACE test set, the accuracy of the models ranged between about 0.6 and 0.85, with RoBERTa-large achieving the highest performance (Table \ref{table1}).

\subsection{Screening for Magnet Options}
\label{section4.2}
\begin{figure}[tbp]
    \begin{center}
        \begin{minipage}[t]{1\linewidth}
            \includegraphics[width=1\linewidth]{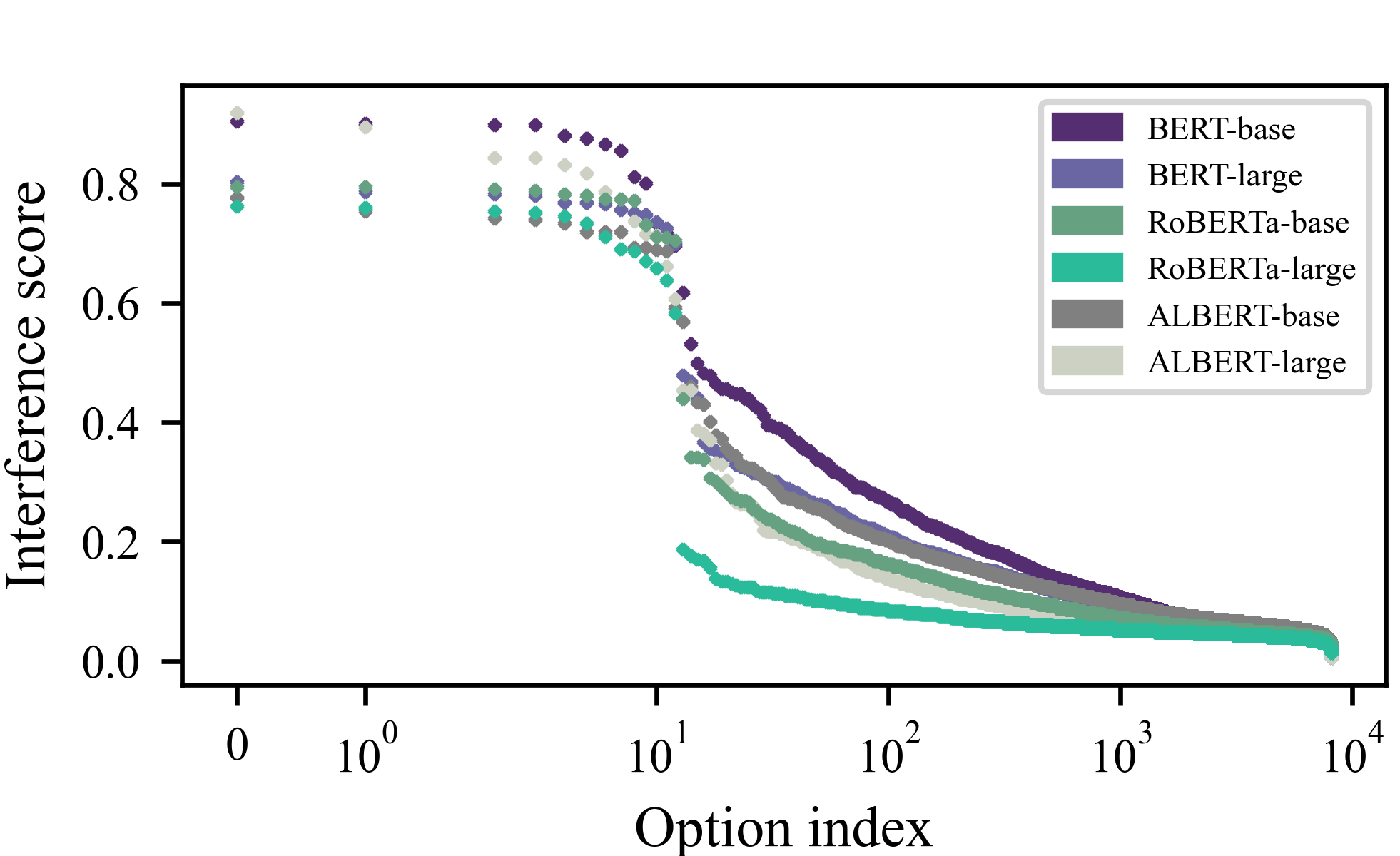}
        \end{minipage}%
        
    \end{center}
    \caption{\label{fig4.2} 
    Interference score evaluated based on a subset of questions.
    }
\end{figure}
The interference score for 8372 options was independently
calculated for each model. Results were shown in Figure \ref{fig4.2},
where the interference score was sorted for each model.
It is found that most of the irrelevant options had a non-zero interference
score, and some irrelevant options yielded high interference scores
around 0.8, which meant the models would choose
those irrelevant options as the answer for about 80\% of the questions. Irrelevant options from the training 
and test sets had similar interference scores 
(Appendix \ref{c-1}).

It was found that the options with exceptionally high interference scores around 0.8 were options that combined other options, such as ``all the above'', which were called the option-combination series. However, not all the magnet options were from the option-combination series. Normal statements, e.g., ``The passage doesn't tell us the end of the story of the movie'', could also reach an average interference score around 0.34.

The correlation between the interference score between models were shown in Appendix \ref{c-2}. We separately showed the results for options from the option-combination series and the others. The correlation coefficient between models had an average value around 0.76, which proved that the interference score was correlated across models. From another perspective, it also implied that our method could work as a model-insensitive adversarial attack method.

\subsection{Validate Magnet Options and Adversarial Attack}
\label{section4.3}
\begin{figure}[tbp]
    \begin{center}
        \begin{minipage}[t]{1\linewidth}
            \includegraphics[width=1\linewidth]{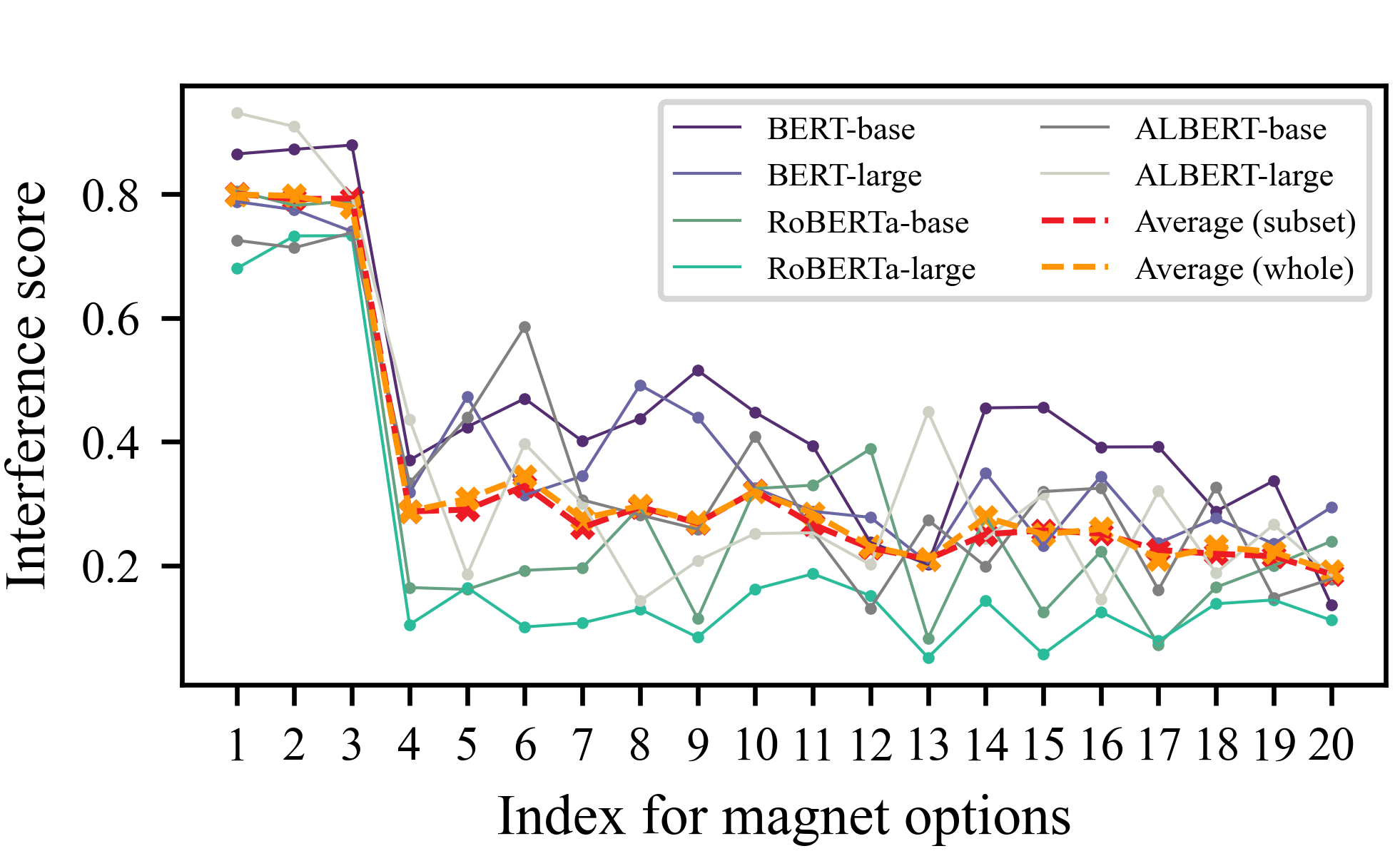}
        \end{minipage}%
        
    \end{center}
    \caption{\label{fig4.3} Interference score 
    evaluated based on the whole RACE test set.}
\end{figure}

We further evaluated the interference score of potential magnet options
based on all the questions in the RACE test set. To construct a set of
magnet options for this analysis, we averaged the interference score
across 3 models, i.e., BERT-large, RoBERTa-large, and ALBERT-large. All
options in \(O_{A}\) were sorted based on the average score, and we
selected 20 options with the highest interference scores to construct
the magnet option set, with the following constraint: Since options with
the highest interference scores were often from the option-combination
series, to increase diversity, we only included 3 options from the
option-combination series. We listed all the 20 magnet options in Appendix \ref{a-2}. The interference score calculated based on the whole RACE
test set was shown in Figure \ref{fig4.3}, which was very similar to the results based on the subset of 346 questions in
Figure \ref{fig4.2}
(comparing average-whole and average-subset in Figure \ref{fig4.3}).

\begin{figure}[tbp]
    \begin{center}
        \begin{minipage}[t]{1\linewidth}
            \includegraphics[width=1\linewidth]{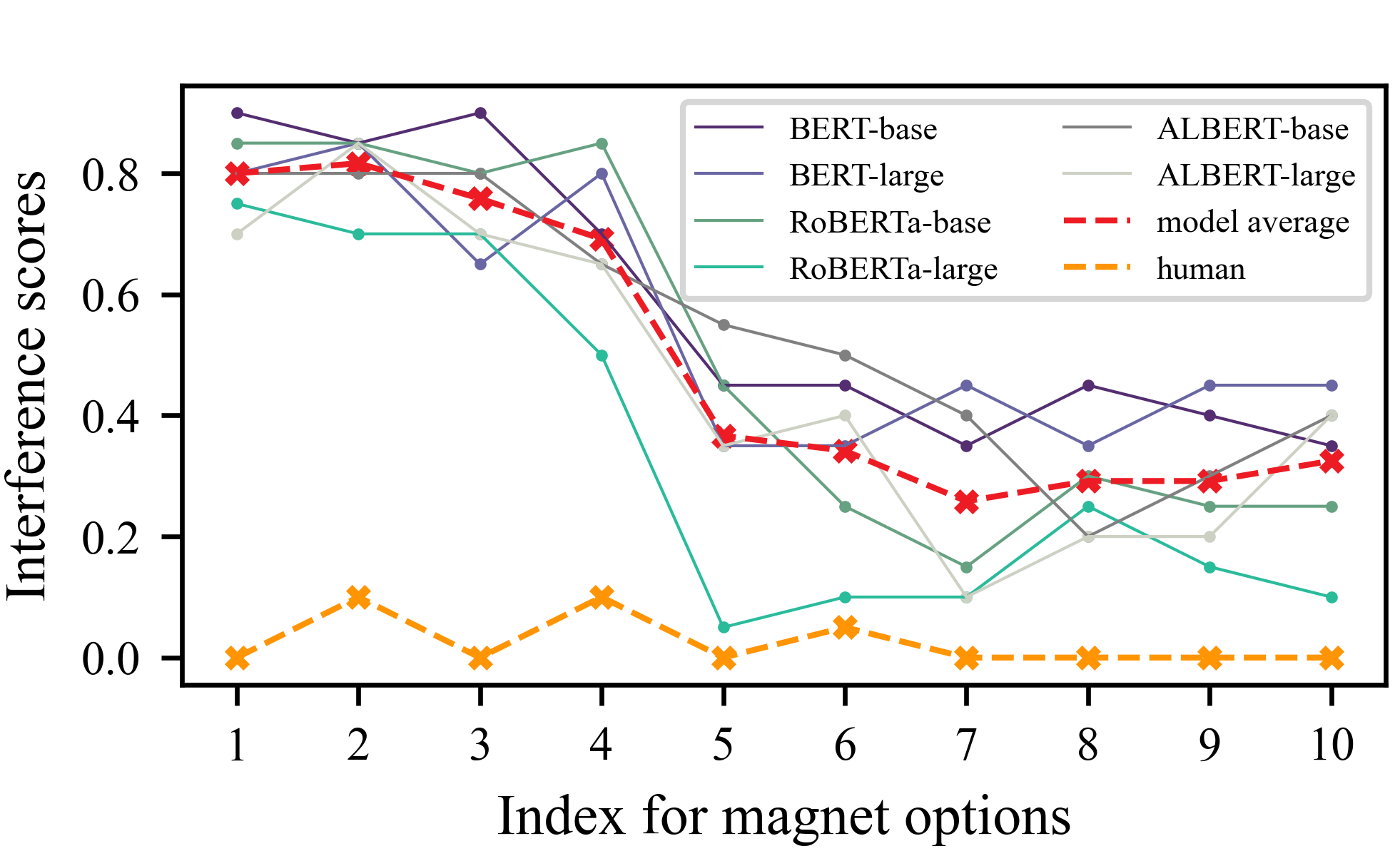}
        \end{minipage}%
        
    \end{center}
    \caption{\label{fig4.4} Interference score for the 
    human experiment and the corresponding 
    interference scores for the models.}
\end{figure}
Table \ref{table1} showed the accuracy of models when attacked by 2 example magnet options. When attacked, the model performance could drop by as much as 0.68.

\begin{table*}[tbp]
    \centering
    \begin{tabular}{ccccccc}
        \hline
    \multicolumn{1}{l}{} & \multicolumn{2}{c}{BERT} & \multicolumn{2}{c}{ALBERT} & \multicolumn{2}{c}{RoBERTa} \\
    Version              & base        & large      & base         & large       & base         & large        \\
    \hline
Original accuracy & 0.614 & 0.681 & 0.683 & 0.752 & 0.738 & 0.846 \\
Adversarial  accuracy$^{1}$     & 0.094 & 0.167 & 0.217 & 0.064 & 0.166 & 0.297 \\
Adversarial  accuracy$^{2}$& 0.381 & 0.524       & 0.334  & 0.506         & 0.656   & 0.798     \\

    \hline
    \end{tabular}
    \caption{\label{table1} 
    Model performance on the RACE test set and model performance after being attacked. The superscript 1 meant use ``A, B and C'' to attack, and the superscript 2 meant use ``The passage doesn't tell us the end of the story of the movie'' to attack.
    }
\end{table*}

\subsection{Human Evaluation}
\label{section4.4}
Next, we verified whether humans were also confused by the magnet options. We randomly selected 20 questions and 10 magnet options. The 10 magnet options selected were listed in Appendix \ref{a-3}. Ten questions were not modified while the other 10 questions were attacked using the procedure shown in Figure \ref{fig3.2}. Twenty human evaluators answered
these 20 questions online. 
The accuracy of humans did not reduce under attack (0.90 in the original samples and 0.94 in the adversarial samples).
The interference score for humans, also the
corresponding interference score for the models, was shown in
Figure \ref{fig4.4}. Humans were not confused by the magnet options.

\subsection{Training with Adversarial Examples}
\label{section4.5}

\begin{table*}[htb]
    \centering
    \begin{tabular}{ccccccc}
    \hline
    base version & BERT                  & ALBERT                 & RoBERTa               \\
    \hline
    Original accuracy & 0.601  & 0.689  & 0.723   \\
    Adversarial  accuracy$^{1}$& 0.576  & 0.681  & 0.725   \\
    Adversarial  accuracy$^{2}$ & 0.670  & 0.740   & 0.778   \\
    \hline
    \end{tabular}
    \caption{\label{table2} 
    Model performance on the RACE test set based on augmented training.
    }
\end{table*}

To reduce sensitivity to magnet options and to potentially reduce the statistical biases of MRC models, we proposed an augmented training method and tested the method using the base version of all models. In the augmented training method, 400 options with the highest interference scores were selected as the irrelevant option set. For each question in the RACE training set, the option set was augmented by adding an option randomly chosen from the irrelevant option set. In other words, although each original question has 4 options, during the augmented training each question has 5 options, including the 4 original options and a randomly chosen irrelevant option. We fine-tuned pre-trained models based on the training set with augmented options.

The accuracy of models fine-tuned using augmented options were shown in Table \ref{table2}, comparable to the original accuracy in Table \ref{table1}. When attacked, however, the accuracy of models fine-tuned using augmented options were much higher than the adversarial accuracy in Table \ref{table1}.

The 1000 options with the highest interference scores were selected to evaluate the effect of augmented training, as shown in Appendix \ref{appendix_b}. Result showed that the interference score dropped for both the 400 options used for augmented training and the other 600 options that were not used for training. Therefore, 
the effect of augmented training could generalize to samples not used for augmented training.

\begin{figure}
    \begin{center}
        \begin{minipage}[]{1\linewidth}
            \includegraphics[width=1\linewidth]{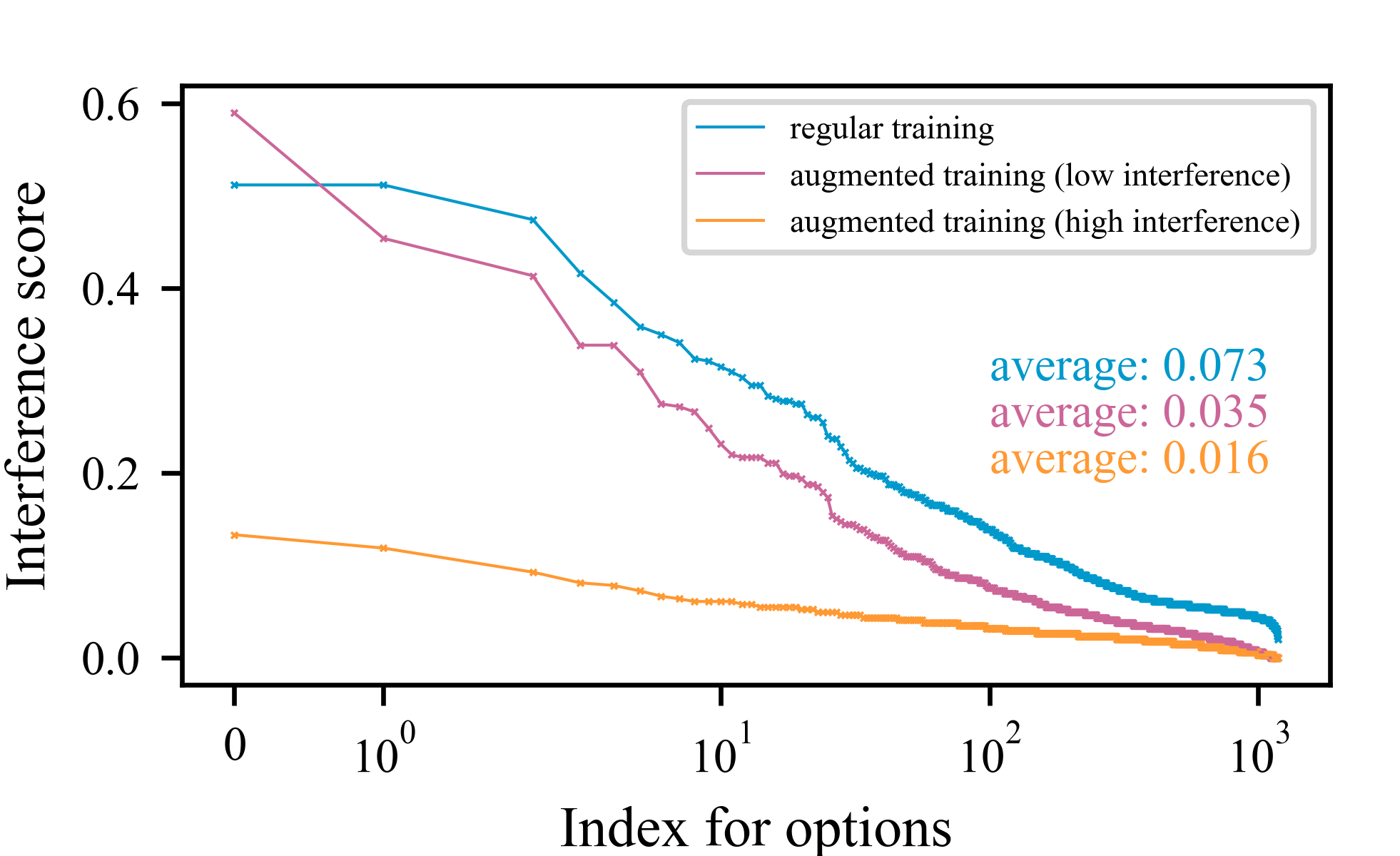}
        \end{minipage}%
        
    \end{center}
    \caption{\label{fig4.5-2} 
    Interference score of 1186 randomly chosen options that are not used in augmented training.
    }
\end{figure}

Another experiment was implemented to explore the impact of irrelevant option set selection. We separately used options with high and low interference scores for training and found that options with higher interference score were more effective at reducing statistical biases (Figure \ref{fig4.5-2}).

\subsection{Interference Score Analysis}
\label{section4.6}
% We did some analyses on the interference score for the same model architecture before and after fine-tuning, the result were shown in Appendix \ref{c-3}. 
Did the statistical biases revealed in previous analyses originate from the pre-training process or the fine-tuning process? Without fine-tuning, the pre-trained models perform poorly on RACE. However, results showed that such an imprecise model can show strong biases (Appendix \ref{c-3}). Interestingly, the interference score is not correlated between the pre-trained model and the fine-tuned model, suggesting that fine-tuning overrides the biases caused by pre-training and introduces new forms of biases.

\section{Related Work}
\label{section5}
Our attack strategy distinguishes from previous work in two ways. First, unlike, e.g., gradient-based methods \citep{ebrahimi-etal-2018-hotflip, DBLP:conf/aaai/ChengYCZH20}, our method does not require any knowledge about the structure of DNN models. Second, some methods manipulate the passage in a passage-dependent way \citep{jia-liang-2017-adversarial,si2020benchmarking,Zhao2018GeneratingNA}, while our method manipulate the options in a passage-independent way. Furthermore, we proposed a strategy to train more robust models that are insensitive to our attack.

Here, we restricted our discussion to RACE, but our method is applicable to other tasks in which the answer is selected from a limited set of options. For example, for span extraction tasks, such as SQuAD, the method will insert a large number of irrelevant phrases into the passage and analyze which phrases are often selected as the answer. In this way, our method is similar to the trigger-based attack methods \citep{wallace-etal-2019-universal}, but the difference is that our method test whether the inserted irrelevant phrase is selected as the answer while the trigger-based methods test whether the content following the trigger phrase is selected.

\section{Conclusion}
\label{section6}
In summary, we propose a new method to evaluate the statistical biases in MRC models. It is found that current MRC models have strong statistical
biases, and are therefore sensitive to adversarial attack. When attacked using the method proposed here, model performance can drop from human-level performance to chance-level performance. To alleviate sensitivity to such attacks, we provided an augmented training procedure that effectively enhances the robustness of models.

\section*{Acknowledgments}
The authors would like to thank the anonymous reviewers for their helpful suggestions and comments. Work supported by Major Scientific Research Project of Zhejiang Lab 2019KB0AC02 and National Natural Science Foundation of China 31771248.

\bibliographystyle{acl_natbib}
\bibliography{acl2021}

\clearpage

\appendix

\section{Experimental Details}
\label{appendix_a}

\subsection{Fine-tuning Parameters}
\label{a-1}
The parameters we used in the process of fine-tuning the pre-trained models were shown in Table \ref{app-table-1}.

\begin{table*}[htbp]
    \centering
        
    \begin{tabular}{ccccccc}
    \hline
    \multicolumn{1}{l}{} & \multicolumn{2}{c}{BERT} & \multicolumn{2}{c}{ALBERT}   & \multicolumn{2}{c}{RoBERTa} \\
        version                   & base & large & base & large & base   & large  \\ \hline
    learning rate        & 1.00E-05   & 1.00E-05    & 2.00E-05     & 1.00E-05      & 1.00E-05  & 1.00E-05        \\
    train epochs         & 5          & 5           & /            & /             & 4         & 4               \\
    train steps          & /          & /           & 12000        & 12000         & /         & /               \\
    train batch size     & 16         & 24          & 32           & 32            & 16        & 16              \\
    warmup steps         & 0          & 0           & 1000         & 1000          & 1200      & 1200            \\
    weight decay          & 0          & 0           & 0            & 0             & 0.1       & 0.1            \\
    \hline
    \end{tabular}

    \caption{\label{app-table-1} 
    Hyperparameters for fine-tuning on RACE. We adapted these hyperparamemers from 
    \citep{lan2019albert,liu2019roberta,ran2019option,zhang2020dcmn+}.
    }
\end{table*}

% fig for test and train b.1
\begin{figure}[tb]
    \begin{center}
        \begin{minipage}[t]{1\linewidth}
            \includegraphics[width=1\linewidth]{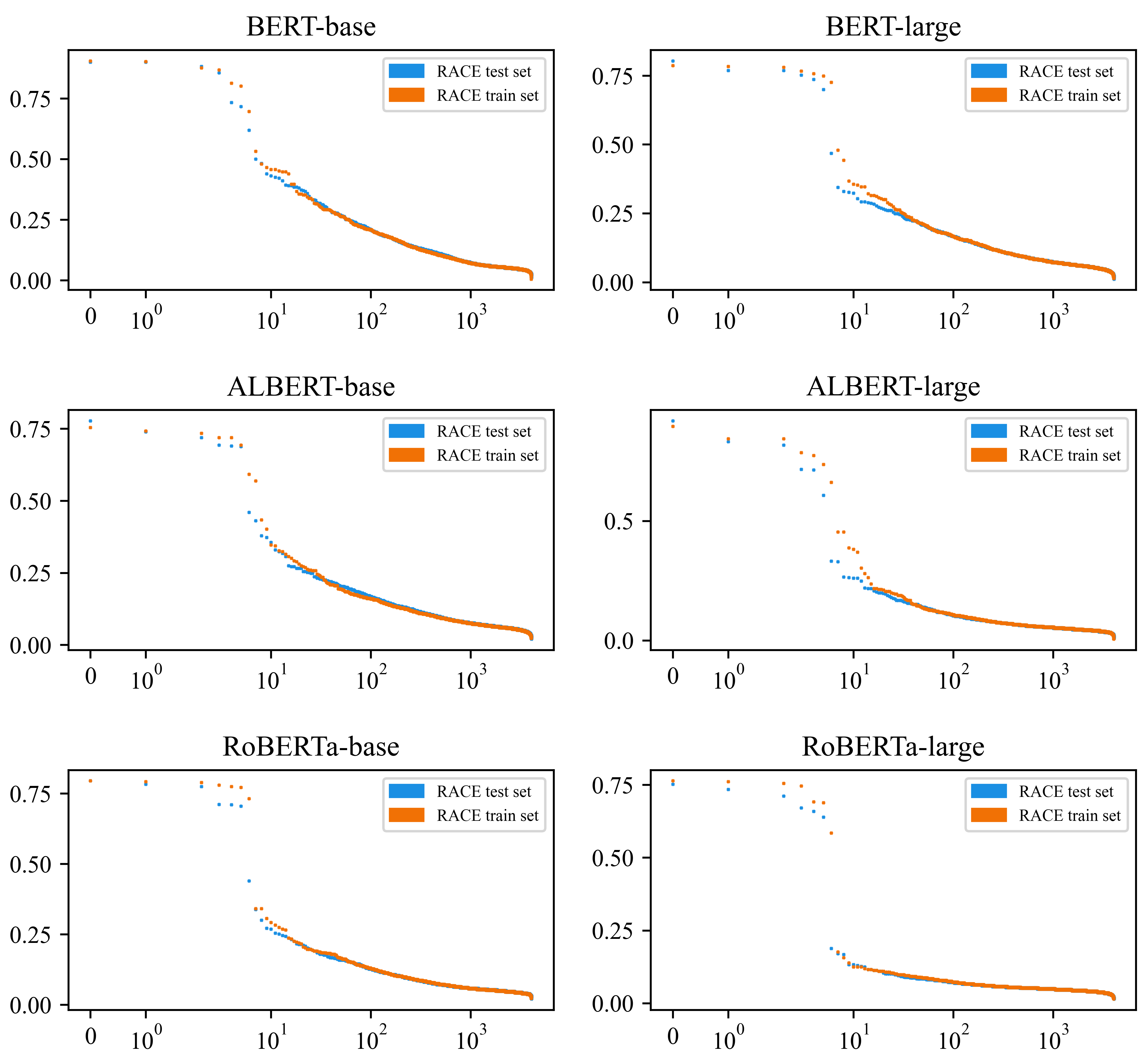}
        \end{minipage}%
        
    \end{center}
    \caption{\label{train_vs_test} 
    Interference score evaluated based on a subset of questions.
    }
\end{figure}

% b.1
\begin{figure}[tbp]
    \begin{center}
        \begin{minipage}[t]{1\linewidth}
            \includegraphics[width=1\linewidth]{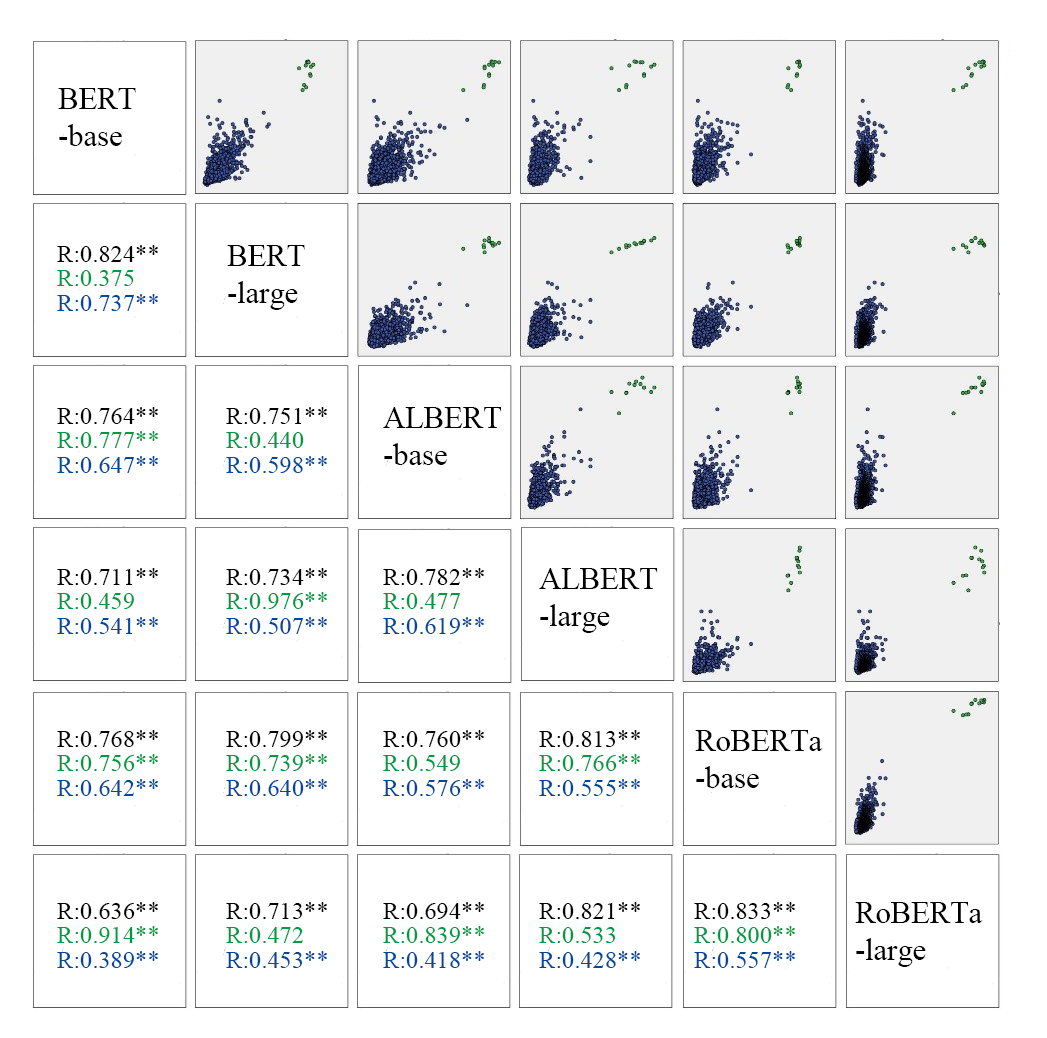}
        \end{minipage}%
        
    \end{center}
    \caption{\label{score-relation} 
    The scatter matrix diagram of the interference scores of the irrelevant options among models.
    }
\end{figure}

\subsection{Magnet Options for Validate}
\label{a-2}
The 20 magnet options used for evaluating the interference scores in Section \ref{section4.3} were shown as following. The sentences selected from the RACE training set were shown in bold.

\begin{enumerate}
    \item A, B and C
    \item \textbf{all of A, B and C}
    \item All of the above.
    \item \textbf{Not all of it can be avoided.}
    \item It's well beyond what the author could be responsible for.
    \item \textbf{The passage doesn't tell us the end of the story of the movie}
    \item didn't give the real answer
    \item \textbf{make us know it's important to listen to people who offer a different perspective through his experience}
    \item \textbf{give us a turning point in mind}
    \item \textbf{not strictly stuck to}
    \item You should purposely go out and make these mistakes so that you can learn from them and not have them ruin your entire life.
    \item what's inside a person is much more important than his/her appearance.
    \item \textbf{Not all of it is man-made Ming dynasty structure.}
    \item \textbf{introduce the topic of the passage}
    \item \textbf{The central command didn't exactly state what had caused the crash.}
    \item \textbf{one good turn deserves another.}
    \item the growing population is not the real cause of the environment problem.,
    \item \textbf{misfortune may be an actual blessing.}
    \item may meet with difficulties sometimes
    \item good answers are always coming when we think outside of the box
\end{enumerate}

\subsection{Magnet Options for Human Evaluation}
\label{a-3}
The 10 magnet options used for human evaluating in Section \ref{section4.4}.
\begin{enumerate}
    \item all the above
    \item Both B and C
    \item do all of the above
    \item A and B
    \item not strictly stuck to
    \item The passage doesn't tell us the end of the story of the movie
    \item It's well beyond what the author could be responsible for.
    \item You should purposely go out and make these mistakes so that you can learn from them and not have them ruin your entire life.
    \item make us know it's important to listen to people who offer a different perspective through his experience
    \item Not all of it is man-made Ming dynasty structure.
\end{enumerate}

\begin{table*}[tbp]
    \centering
    \begin{tabular}{ccccccc}
    \hline
    BERT-base      & Correlation coefficient & accuracy & Average interference score \\\hline
    Pre-trained model & -0.023                                        & 0.315    & 0.0518                     \\
    Partly fine-tuned model & 0.069*                                        & 0.315    & 0.0214                     \\
    Fine-tuned model    & 1                                             & 0.613    & 0.0713                     \\\hline
    RoBERTa-base   & Correlation coefficient& accuracy & Average interference score \\\hline
    Pre-trained model & -0.021                                        & 0.225    & 0.3553                     \\
    Partly fine-tuned model & 0.088**                                       & 0.289    & 0.2282                     \\
    Fine-tuned model    & 1                                             & 0.743    & 0.0569                     \\\hline
    ALBERT-base    & Correlation coefficient& accuracy & Average interference score \\\hline
    Pre-trained model & -0.013                                        & 0.254    & 0.1483                     \\
    Partly fine-tuned model & 0.231**                                       & 0.39     & 0.1043                     \\
    Fine-tuned model    & 1                                             & 0.702    & 0.0703          \\          
    \hline
\end{tabular}
\caption{\label{app-table-2} 
Interference score of 1000 randomly selected irrelevant options for the same model architecture before and after fine-tuning. Correlation coefficient was counted between the interference score before and after fine-tuning (** $p<0.01$, and * $p<0.05$).
}
\end{table*}
\begin{figure}
    \begin{center}
        \begin{minipage}[H]{0.9\linewidth}
            \includegraphics[width=1\linewidth]{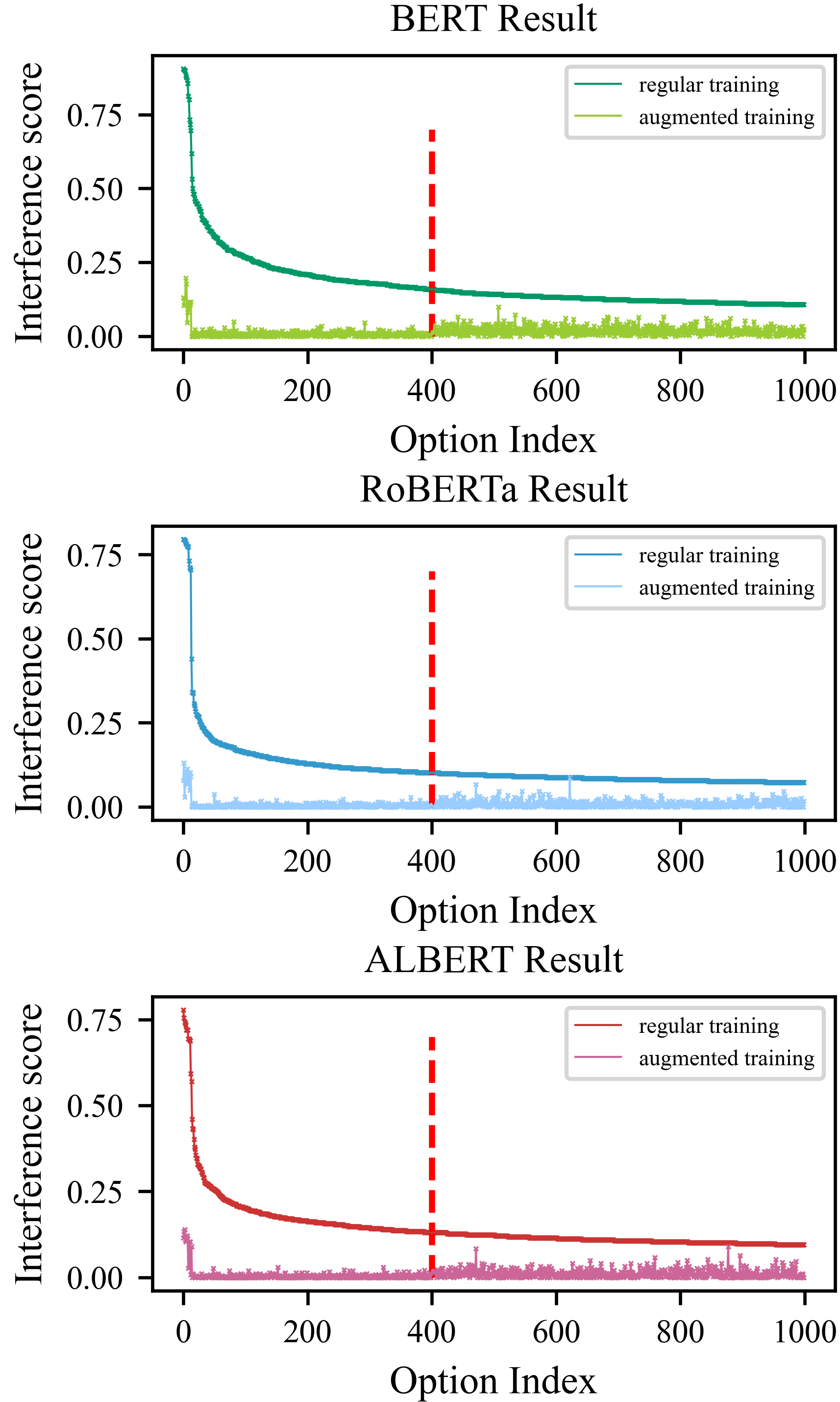}
        \end{minipage}
        
    \end{center}
    \caption{\label{fig4.5-1} 
    Interference score comparison of models evaluated based on a subset of questions.
    }
\end{figure}
\section{Study of Interference Score}
\label{appendix_c}

\subsection{Comparison of Irrelevant Options from RACE Training and Test Set}
\label{c-1}
Different models in Figure \ref{fig4.2} were separately shown in Figure \ref{train_vs_test}. It denoted that irrelevant options from the training and test sets had similar interference score. Only in BERT-large and ALBERT-large models, the interference scores of the irrelevant options from the training set were higher than those from the test set in a certain range.

\subsection{Comparison of Interference Scores Based on Different Models}
\label{c-2}
The scatter matrix diagram of the interference scores of the irrelevant options among different models was shown in Figure \ref{score-relation}. The detailed experimental process was described in Section \ref{section4.2}. Here, text in black showed the correlation coefficient of all options; text in green showed the options of the option-combination series; text in blue showed the options except the option-combination series.

In general, the interference scores between models had high correlation coefficients. Models from the same architecture were more likely to have similar interference scores.

\subsection{Comparison of Interference Scores During Fine-tuning}
\label{c-3}
For each model architecture, the pre-trained model, partly fine-tuned model (fine-tuned the linear transformation mentioned in Section \ref{section2}), and fully fine-tuned model were collected, and were used to evaluate the interference score of 1,000 randomly selected irrelevant options. The results were shown in Table \ref{app-table-2}. The subset of questions mentioned in Section \ref{section4.1} were used to evaluate the interference score.

\section{Augmented Training Result}
\label{appendix_b}
The augmented training results were shown in Figure \ref{fig4.5-1}. In the figures, the left side of the red line contains the irrelevant options that were used in augmented  training, and the right is the irrelevant options that were not involved in augmented  training.

\end{document}